\documentclass[letterpaper, 10 pt, journal, twoside]{ieeetran}

\IEEEoverridecommandlockouts                              
\usepackage[T1]{fontenc}

\usepackage[noadjust]{cite}

\usepackage{filecontents}
\usepackage{amsmath}
\DeclareMathSymbol{:}{\mathord}{operators}{"3A}  
\usepackage{yhmath} 
\usepackage{mathtools} 
\usepackage{amssymb} 
\usepackage{xfrac} 
\usepackage{sidecap}
\usepackage{graphicx}
\usepackage{subcaption}
\usepackage{multirow} 
\usepackage{fancyhdr}
\usepackage{leftidx} 
\usepackage{cancel} 
\usepackage{caption}
\usepackage[super]{nth} 
\usepackage[normalem]{ulem} 
\usepackage{color}
\usepackage{xcolor} 
\usepackage{booktabs} 
\usepackage{siunitx} 
\usepackage{booktabs} 
\usepackage{colortbl} 
\usepackage{textcomp} 
\usepackage{tikz}
\usepackage[noadjust]{cite} 
\usepackage{xcolor}
\usepackage{wasysym}  

\newcommand{\RNum}[1]{\uppercase\expandafter{\romannumeral #1\relax}} 

\newcommand{\realfield}[1]{\hbox{I \kern -.5em R}^{#1}}
\newcommand {\mb}[1]{\mathbf{#1}}
\newcommand {\bs}[1]{\boldsymbol{#1}}
\newcommand{\uvec}[1]{\hat{\mathbf{#1}}}
\newcommand{\Rot}[2]{{^{#1}\mathbf{R}}_{#2}}  
\newcommand{\T}{^{\mathrm{T}}}  

\newcommand*\circled[1]{\tikz[baseline=(char.base)]{\node[circle,minimum size=7pt,draw=black,inner sep=0.5pt](char){\scriptsize #1};}}

\usepackage{enumerate}

\newif\ifTrackChanges   
\TrackChangestrue 
\definecolor{deep-red}{RGB}{192, 0, 0}
\definecolor{deep-purple}{RGB}{120, 0, 170}
\definecolor{good-green}{RGB}{0,175,0} 
\definecolor{purple}{RGB}{210, 0, 210} 
\definecolor{alizarin}{rgb}{0.82, 0.1, 0.26}

\usepackage[textsize=tiny, bordercolor=white,backgroundcolor=gray!30,linecolor=black,colorinlistoftodos]{todonotes}  
\usepackage[normalem]{ulem} 
\usepackage{soul} 

\usepackage[mathscr]{euscript}
\ifTrackChanges

    \newcommand{\cut}[1]{{\color{gray}{#1}}}
\else

    \newcommand{\cut}[1]{{}}

\fi

\usepackage{algorithm,algorithmicx}
\usepackage[noend]{algpseudocode}

\algnewcommand\algorithmicinput{\textbf{Input:}}
\algnewcommand\Input{\item[\algorithmicinput]}

\algnewcommand\algorithmicoutput{\textbf{Output:}}
\algnewcommand\Output{\item[\algorithmicoutput]} 

\graphicspath{{figures/pdf/}}

\newbox\tempbox
\makeatletter
\let\NAT@parse\undefined
\makeatother

\usepackage{hyperref} 
\hypersetup{
  colorlinks   = true, 
  urlcolor     = blue, 
  linkcolor    = blue, 
  citecolor   = green 
}

\usepackage{balance}
\title{Fluoroscopic Shape and Pose Tracking of Catheters \\
with Custom Radiopaque Markers}

\author{Jared Lawson$^{1}$, Rohan Chitale$^{2}$ and Nabil~Simaan$^{1}$$^{\dag}$
\thanks{Manuscript received: February 7, 2025; Revised May 16, 2025; Accepted June 10, 2025.}
\thanks{This paper was recommended for publication by Editor Jessica Burgner-Kahrs upon evaluation of the Associate Editor and Reviewers' comments. J. Lawson was supported in part by NIH award \#T32EB021937 and by Vanderbilt University funds.}
\thanks{$\dag$ Corresponding author}
\thanks{$^{1}$Department of Mechanical Engineering, Vanderbilt University, Nashville, TN 37235, USA
        {\tt\small (jared.p.lawson, nabil.simaan) @vanderbilt.edu}}%
\thanks{$^{2}$Department of Neurological Surgery, Vanderbilt University Medical Center, Nashville, TN 37235, USA
        {\tt\small (rohan.chitale) @vumc.org}}%
\thanks{Digital Object Identifier (DOI): see top of this page.}
}
\begin{document}
\maketitle


\markboth{IEEE Robotics and Automation Letters. Preprint Version. June, 2025}{Lawson \MakeLowercase{\textit{et al.}}: Fluoroscopic Shape and Pose Tracking of Catheters with Custom Radiopaque Markers} 

\thispagestyle{fancy}
\fancyhf{}
\renewcommand{\headrulewidth}{0pt}
\lhead{IEEE Robotics and Automation Letters. Preprint Version. June, 2025}
\rfoot{\centering \scriptsize \copyright 2025 IEEE. Personal use of this material is permitted. Permission from IEEE must be obtained for all other uses, in any current or future media, including reprinting/republishing this material for advertising or promotional purposes, creating new collective works, for resale or redistribution to servers or lists, or reuse of any copyrighted component of this work in other works.}


\begin{abstract}
Safe navigation of steerable and robotic catheters in the cerebral vasculature requires awareness of the catheter's shape and pose. Currently, a significant perception burden is placed on interventionalists to mentally reconstruct and predict catheter motions from biplane fluoroscopy images. Efforts to track these catheters are limited to planar segmentation or bulky sensing instrumentation, which are incompatible with microcatheters used in neurointervention. In this work, a catheter is equipped with custom radiopaque markers arranged to enable simultaneous shape and pose estimation under biplane fluoroscopy. A design measure is proposed to guide the arrangement of these markers to minimize sensitivity to marker tracking uncertainty. This approach was deployed for microcatheters ($\leq \diameter2\textrm{mm}$) navigating phantom vasculature with shape tracking errors <1mm and catheter roll errors $<40^{\circ}$. This work can enable steerable catheters to autonomously navigate under biplane imaging.
\end{abstract}

\section{Introduction} \label{sec:intro}
\par Endovascular neurosurgery is a rapidly growing domain which enables treatment of cerebrovascular disease with minimally-invasive approaches. Among the most common endovascular neurointerventions include aneurysm coiling and mechanical thrombectomy (MT), which has become the gold standard for treating strokes caused by large vessel occlusions (LVOs).
\par While these treatment options offer the potential to save and improve quality of life for many patients, they are only performed by a limited subset of highly-trained neurosurgeons, neurologists and interventional radiologists \cite{Tripathi2023}. The limited number of capable interventionalists is in large part due to the technically demanding nature of catheter navigation, especially in the distal regions of the cerebrovasculature, where blood vessels are small, delicate, and highly tortuous. The high level of tortuosity requires that interventionalists use biplane fluoroscopy to simultaneously view the anatomy and their devices from multiple views, demanding of them a significant perception burden during the procedure.
\par A standard biplane fluoroscopy setup for neuronavigation includes an anterior-posterior (AP) and lateral view to capture the three-dimensional (3D) intricacies and curvatures of the patient’s vasculature. While these disparate views capture planar bending features, the interventionalist must rely on a mental reconstruction of the two views to determine how the anatomy, and their devices, interact in 3D space. 
\par This challenge is exacerbated as actively steerable catheters are developed for endovascular navigation. Unlike passive catheters, which are typically circumferentially symmetric, steerable catheters are designed to bend in a specific plane. Control of these steerable catheters demands the ability to track the complete their pose and shape.
\par
\begin{figure}[htbp]
        \centering
        \includegraphics[width=0.95\columnwidth]{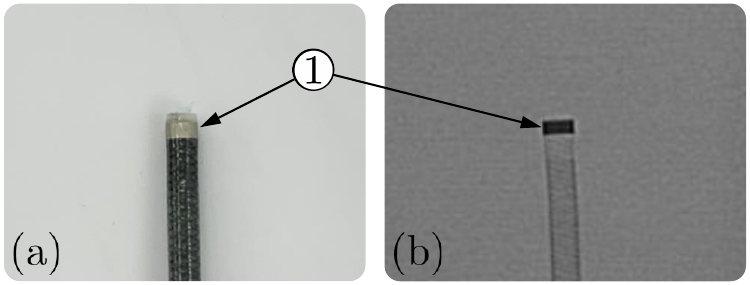}
        \caption{An example of radiopaque markers embedded on catheters for fluoroscopic guidance. (a) A Medtronic React 71 catheter under camera image, while (b) shows the same catheter under fluoroscopy, with \protect\circled{1} highlighting the radiopaque marker band at the catheter's distal tip.}\label{fig:radiopaque_tip}
\end{figure}

 While larger scale catheters have been equipped with extrinsic pose sensing, such as electromagnetic (EM) tracking, the microcatheters used at the scale of neuronavigation are too small for embedding sensor coils, and may also require shape information rather than tip pose alone \cite{Schwein2018em}. The catheters for cerebral vasculature access are less than 2mm in diameter with maximized inner lumens to retain good clot aspiration abilities. Therefore, most catheters have wall thicknesses of $\sim0.2$mm (0.008"), making it difficult to embed sensing. Segmentation-based approaches have been shown to accurately localize catheter tip positions under fluoroscopy by embedding radiopaque marker bands either at the tip (see Fig.~\ref{fig:radiopaque_tip}) or at multiple locations along the catheter length, but they have not been used to accurately determine the complete orientation of the catheter tip \cite{Hwang2018}.
\par The contribution of this paper is the development of a catheter shape and pose sensing approach utilizing biplane fluoroscopy. The key aspects of this main contribution include the design and kinematics of a steerable catheter tip equipped with discrete radiopaque markers, the segmentation and shape sensing algorithm itself, as well as a design optimization framework. This framework analyzes how the catheter geometry and placement of markers affects the sensing algorithm. The feasibility of these approaches are demonstrated through simulation studies, as well as experiments of a catheter prototype under fluoroscopy.
\par In the following section, the relevant state-of-the-art is presented, highlighting the need for full shape and pose estimation of small catheters under biplane fluoroscopy. In Section~\ref{sec:kinematics}, the kinematics of the catheter shape are presented, as well as the design of the radiopaque markers. Section~\ref{sec:estimationapproach} includes the methods comprising the core contribution of this work, which incorporates image segmentation, biplane reconstruction of marker positions, and finally shape and pose estimation. The design of the radiopaque marker placement is studied in Section~\ref{sec:designOpt}, resulting in a formulation for optimal marker spacing for a given catheter and imaging system. The performance of this estimation approach is evaluated in fluoroscopic experiments in Section~\ref{sec:eval}, before these results and conclusions are discussed in Section~\ref{sec:conclusion}.
%
%
%
%
%
\section{Related Works} \label{sec:related_works}
%
%
\par The two principal approaches for shape and pose estimation of catheters include leveraging sensor feedback or image feedback, and in some cases a hybrid of both. The state-of-the-art techniques have been recently reviewed in \cite{Ramadani2022survey}.
\par Extrinsic shape or pose estimation involves the placement of sensors along the catheter itself. While this refers to any method of external sensing, the most common modalities include embedding electromagnetic (EM) coils \cite{Mefleh2014,Song2015em,Schwein2018em,Ha2023sensor}. While EM sensing can offer good spatial resolution at one point on the catheter, many coils would be needed to estimate shape information.
%
%
Optical fibers with fiber Bragg gratings (FBGs) are a relatively low-profile sensing alternative that enable strain sensing along the length of the catheter and for solving shape and pose \cite{Liu2015shape,Roesthuis2013,Ourak2021fusion}.
The limitations to using FBGs include their sensitivity to thermal effects and their cost. To account for thermal effects, previous works embed a sensor along the central axis, however for hollow lumen catheters, this is unachievable. Both EM and FBG solutions additionally provide no information relating the catheter to the anatomy, which is critical for safe use of the devices.
\par Shape and pose estimation from the biplane fluoroscopy images represents an approach intrinsic to the clinical workflow. In the independent image planes, both traditional \cite{Fazlali2015,Chang2016,Omisore2021} 
and machine learning based segmentation methods
\cite{Torabinia2021DL,Hashemi2024}
have been used to find the shape of the planar projection of the catheter or guidewire.  Reconstruction of the biplanar segmentations relies on projective geometry which can leverage either a series of discrete points
\cite{Bender1999recon,Brost2009,Baur2016recon}
or segmented splines 
\cite{Hoffmann2013recon,Delmas2015,Wagner20164d}. 
Amongst these approaches, however, they are unable to capture rotation about the axis of the catheter, since these devices are circumferentially symmetric. Two disparate methods proposed image segmentation approaches which leverage single-plane fluoroscopy, but assume that the catheter bends in a circular fashion which is not a strong assumption for most catheters \cite{Hwang2018,ravigopal2021}. Roll estimation was demonstrated for intracardiac echocardiography (ICE), however this doesn't additionally capture the catheter shape \cite{Annabestani2023roll}. Finally, our previous work determined a steerable catheter's bending plane and pose, however it assumed that bending occurs in one plane, and that the amount of bending is estimated from robotic steering \cite{Lawson2023est}. This work improves this method by simultaneously solving both shape and pose, and not limiting the shape to an assumed bending mode.
\section{Catheter Design and Kinematics}\label{sec:kinematics}
\par In this section, we discuss the kinematics of the distal segment of a flexible catheter. First we describe the shape of the backbone curve of the catheter in Section~\ref{sec:cath_kinematics}. In Section~\ref{sec:estimation_helical_kinematics}, the distribution of radiopaque markers along the catheter's outer surface are defined.
%
\subsection{Catheter Kinematics}\label{sec:cath_kinematics}
%
%
\par The distal segment of the catheter is treated as an elastic rod that is subject to bending strain and neglects torsion, shear, and extension. Although there can be a buildup of torsion along the length of a catheter, we only consider the distal steerable tip which observes relatively negligible torsion compared to the full catheter length.
\par
\begin{figure}[htbp]
        \centering
        \includegraphics[width=\columnwidth]{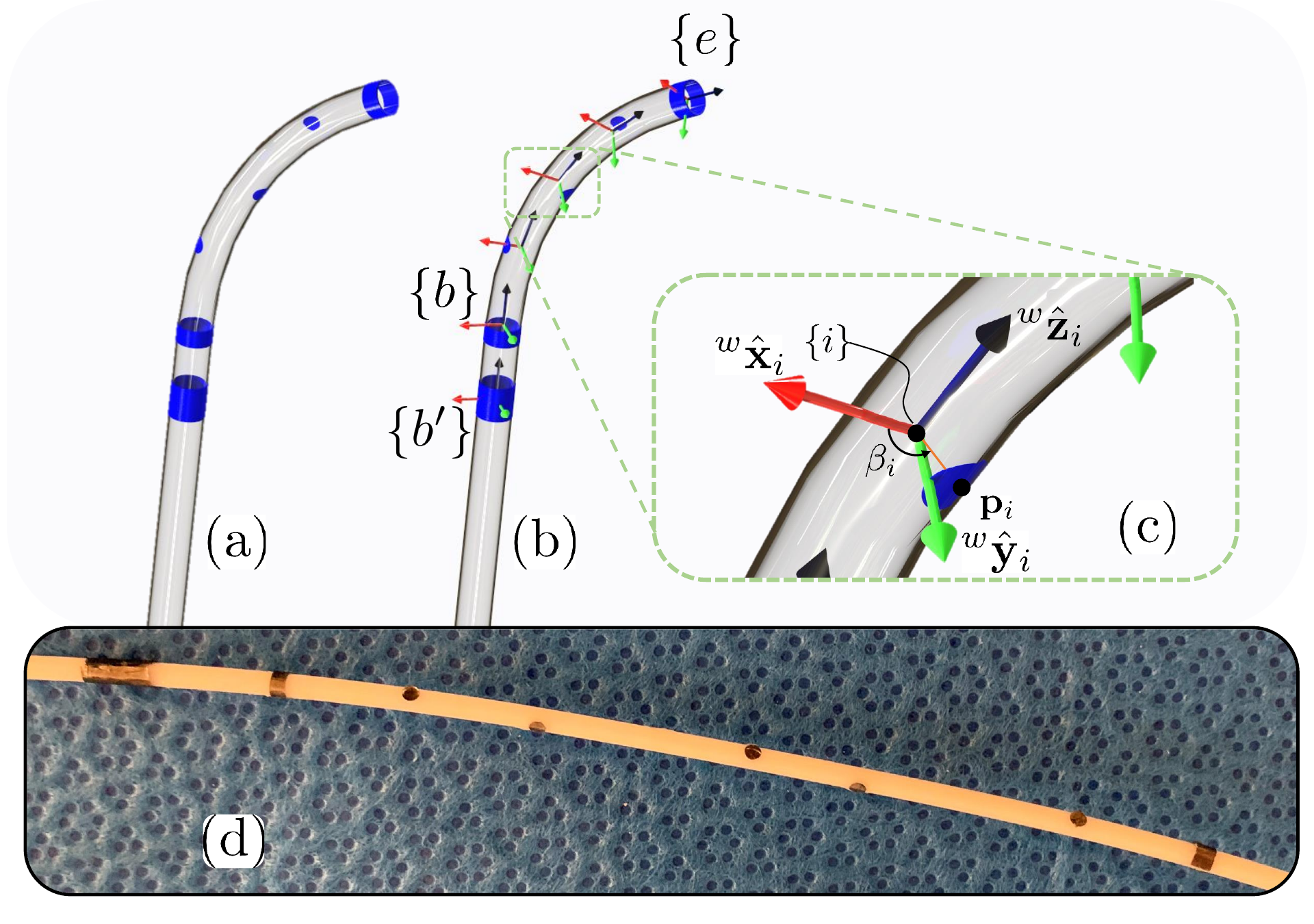}
        \caption{The catheter tip with radiopaque markers. (a) a catheter design with the catheter shown as transparent, and the radiopaque markers shown in blue. (b) the same configuration with frames assigned along the catheter's central axis. (c) an inset of an individual frame, indicating the placement of the marker in the x-y plane of local frame, $\{i\}$. (d) includes physical prototype of the markers on Pebax tubing.}\label{fig:catheter_frames}
\end{figure}

 The portion of the catheter tracked under fluoroscopy includes its distal tip, which is prescribed with frame $\{e\}$, along with a segment of length $L$ from the tip frame which terminates at a base frame, prescribed $\{b\}$. The backbone curve, $\mb{p}(s)$, along the catheter's central axis from $\{b\}$ to $\{e\}$ is parameterized by arc-length variable, $s\in\left[0,L\right]$, where $s=0$ at $\{b\}$ and $s=L$ at $\{e\}$ (see Fig.~\ref{fig:catheter_frames}). The bending curvature along the backbone curve defines a local-frame distribution of strain $\mb{u}(s)$. Orekhov et. al. \cite{Orekhov2023shapesense} used a modal representation of the strain distribution such that:
\begin{equation}\label{eq:strain_modal1}
    \mb{u}(s,\mb{c}) = \bs{\Phi}(s)\,\mb{c}. 
\end{equation}
where $\mb{c}\in\realfield{3m}$ is a vector of modal coefficients, $\bs{\Phi}(s)\in\realfield{3\times 3m}$ is a matrix with modal basis functions, and $m$ refers to the order of the basis function.  The components of strain can be further decomposed:
\begin{equation}\label{eq:strain_modal2}
    \begin{bmatrix}
        u_x(s,\mb{c}) \\
        u_y(s,\mb{c}) \\
        u_z(s,\mb{c})
    \end{bmatrix} = \begin{bmatrix}
        \bs{\phi}_{x}\T(s) & \mb{0} & \mb{0} \\
        \mb{0} & \bs{\phi}_{y}\T(s) & \mb{0} \\
        \mb{0} & \mb{0} & \bs{\phi}_{z}\T(s)
    \end{bmatrix}\begin{bmatrix}
        \mb{c}_{x} \\ \mb{c}_{y} \\ \mb{c}_{z}
    \end{bmatrix}
\end{equation}
\par\noindent where $\bs{\phi}(s)$ are vectors of modal basis functions and $\mb{c}_{x}$, $\mb{c}_{y}$, and $\mb{c}_{z}$ are the modal coefficients for each bending direction. The choice of modal basis functions includes monomial bases, Legendre polynomials, and Chebyshev polynomials \cite{Orekhov2023shapesense}. In this work, we use Chebyshev polynomials of the first kind
since they form an orthogonal basis with normalized values between $\left[-1,1\right]$ - therefore providing good numerical conditioning for the modal factors identification problem and minimizing Runge's phenomenon \cite{mason2002chebyshev}.
\par Per our assumption of negligible torsional strain, the z-terms in Eq.~\eqref{eq:strain_modal2} vanish. Therefore, \mbox{$u_z(s,\mb{c})=0$} and $\mb{c}=\left[\mb{c}_x\T,\mb{c}_y\T\right]\T\in\realfield{2m}$ for the remainder of this paper.
\par Given the strain distribution $\mb{u}(s,\mb{c})$ in the local frame \{T\}, the backbone curve can be obtained by Lie group integration of the following differential equation \cite{selig2007geometric}:
\begin{equation}\label{eq:dTds_1}
    \frac{\partial \mb{T}(s,\mb{c})}{\partial s} = \mb{T}'(s,\mb{c}) = \mb{T}(s,\mb{c})\widehat{\bs{\xi}}(s,\mb{c})
\end{equation}
where $\mb{T}(s)\in SE(3)$ is the homogeneous transformation of section-attached frame \{T\} as shown in Fig.~\ref{fig:catheter_frames}, and $\widehat{\bs{\xi}}(s)\in \mathfrak{se}(3)$ is given by:
\begin{equation}
    \widehat{\bs{\xi}}(s) = \begin{bmatrix}
        0 & -u_{z}(s) & u_{y}(s) & 0 \\
        u_{z}(s) & 0 & -u_{x}(s) & 0 \\
        -u_{y}(s) & u_{x}(s) & 0 & 1 \\
        0 & 0 & 0 & 0
    \end{bmatrix}
\end{equation}
\par Assuming the pose of the catheter base is known at the base frame, $\mb{T}(0)$, the pose of any frame, $\mb{T}(s_i)$, along the catheter's central axis can be found using a moving frame product-of-exponentials sequence \cite{selig2007geometric}:
%

\begin{equation}\label{eq:prodexpDK}
    \mb{T}(s_{i},\mb{c}) = \mb{T}(0)\prod_{j=1}^i e^{\Delta s\, \widehat{\bs{\xi}}(s_{(j-1)},\mb{c})} 
\end{equation}
\par \noindent where $\Delta s$ is the step size along the arc-length.
%
\subsection{Marker Design Kinematics}\label{sec:estimation_helical_kinematics}
\par The design of the catheter tip incorporates discrete radiopaque markers placed at known points along the catheter's surface (see Figure~\ref{fig:catheter_frames}). The tip frame, $\{e\}$, has a round marker band surrounding the diameter of the catheter, which is standard for most commercially available catheters. At the base of the segment are two round marker bands: one placed at the base frame, $\{b\}$, and another placed proximally labeled $\{b'\}$. With these two markers, the local tangent axis at the base frame can be computed to give an estimate of $\mb{T}(0)$.
\par If only round marker bands surrounding the the catheter's diameter were used, the catheter shape could be found but the pose would not be discernible as each marker would be symmetric about the catheter backbone curve. Therefore, the remaining markers are placed on the catheter's surface in a generalized helical pattern about the backbone curve, $\mb{p}(s)$. For example, the placement of the $i^{th}$ marker can be defined by it's arc-length, $s_{i}$, and it's angle, $\beta_{i}$. Given the backbone curve, $\mb{p}(s,\mb{c})$, the position of the $i^{th}$ marker is:
\begin{subequations}
    \begin{equation}\label{eq:dk_marker_1}
        \mb{T}(s_i,\mb{c}) = \begin{bmatrix}
            \mb{R}(s_{i},\mb{c}) & \mb{p}(s_{i},\mb{c}) \\
            \mb{0} & 1
        \end{bmatrix}\in\textrm{SE}(3)
    \end{equation}
    \begin{equation}\label{eq:dk_marker_2}
        \mb{p}_{i}(s_i,\mb{c}) = \mb{p}(s_{i},\mb{c}) + \mb{R}(s_{i},\mb{c})\begin{bmatrix}
        r\cos(\beta_{i}) \\
        r\sin(\beta_{i}) \\
        0
    \end{bmatrix}
    \end{equation}
\end{subequations}
\par \noindent where $r$ is the radius of the catheter's cross section. While in this work we will assume a fixed radius, this radius could be a function of arc-length, $\mb{r}(s)$, to generalize to variable radius devices. Eq.~\eqref{eq:dk_marker_1} computes the pose of the backbone at the marker's arc-length, $s_{i}$, and Eq.~\eqref{eq:dk_marker_2} includes the radial offset to the catheter's surface from the backbone frame.
\par This parameterization enables a variable-shape marker layout that can incorporate straight, helical, and variable-pitch helical patterns among others. In total, a given catheter will include $n$ markers (excluding the base and tip markers), with design parameters of arc-lengths, $\mb{s}_{d} = \left[s_{1},\dots,s_{n}\right]\T$ and angles, $\bs{\beta}_{d} = \left[\beta_{1},\dots,\beta_{n}\right]\T$. The impact of these design parameters will be investigated further in Section~\ref{sec:designOpt}.

\section{Shape and Pose Estimation}\label{sec:estimationapproach}
\begin{figure*}[htbp]
        \centering
        \includegraphics[width=0.9\textwidth]{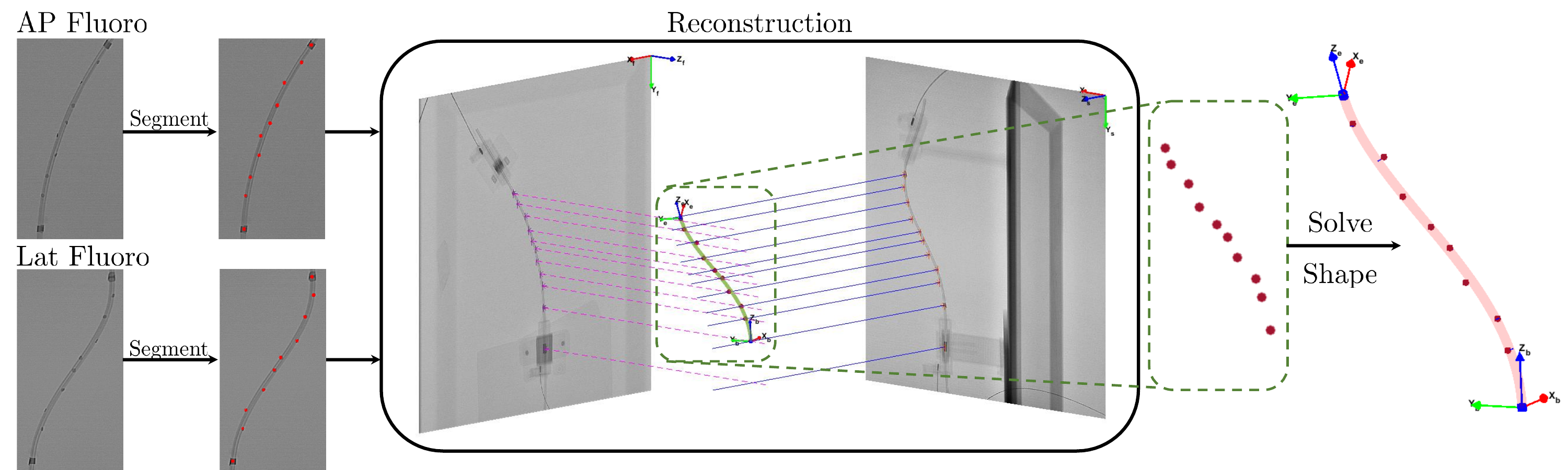}
        \caption{A sample overview of the estimation procedure from input AP and Lateral fluoroscopic images (left) through segmentation, reconstruction, and shape estimation, respectively.}\label{fig:estimation_workflow}
\end{figure*}
\par With knowledge of how to represent the catheter kinematics, as well as a design for the location of radiopaque markers, we look to extract both the shape and pose of the catheter from the available clinical imaging modality, biplane fluoroscopy. This section describes the overview of the estimation approach consisting of two algorithms. The first algorithm consists of segmenting the planar locations of the radiopaque markers in each image plane. Once segmented, the second algorithm reconstructs the marker 3D positions and solves for the modal shape coefficients $\mb{c}$ and base frame rotation about the backbone, $\sigma$, resulting in complete shape and pose information. An overview of this is shown in Fig.~\ref{fig:estimation_workflow}.
\subsection{Biplanar Image Segmentation}\label{sec:segmentation}
\par In both the AP and lateral planes, radiopaque markers appear as dark groupings of pixels that are segmented to determine the planar positions of the markers. In this section, the segmentation process is described for an individual image, however it is performed on both planes. These images will be assumed to be calibrated by the fluoroscopy system with known pixel to millimeter conversions. 
\par Algorithm~\ref{alg:segmentation} starts with subtracting the  background of the image to highlight prominent features, which are then thresholded to identify pixels matching the intensity of the radiopaque marker. The returned noisy result is then used to compute maximally stable extremal regions (MSER) which group likely marker candidate pixels using MATLAB's built-in functionality, $\texttt{detectMSERfeatures}\left(\right)$ \cite{Nister2008linear}. The centroid locations of each of these elements give planar positions of each marker in the image frame (Line 4: $\texttt{GetCentroids}$). Finally, the MSER features are filtered by area (Line 5: $\texttt{GetAreas}$) to determine which markers correspond to the base and tip of the catheter. 
\par This segmentation algorithm, which is implemented in MATLAB 2021a, can analyze an image at $\sim40$Hz, which is faster than the fluoroscopy images update ($\sim10$Hz). For each frame update, the base markers (${^{f,s}\mb{p}_{b,b'}}$), tip markers (${^{f,s}\mb{p}_{e}}$), and an unordered set of the intermediate markers (${^{f,s}\tilde{\mb{p}}_{u,int}}$) are found, where $\{f\}$ and $\{s\}$ refer the front (AP) and side (lateral) image frames.
%
\begin{algorithm}
\caption{Radiopaque Marker Segmentation}\label{alg:segmentation}
\begin{algorithmic}[1]
\Require img, threshParam, bgIntensity, areaRange
\State imgSub = img - bgIntensity
\State imgThresh = Thresholding(imgSub, threshParam)
\State MSER = detectMSER(imgThresh, areaRange)
\State centroids = GetCentroids(MSER)
\State areas = GetAreas(MSER)
\end{algorithmic}
\end{algorithm}
%
%
\subsection{Marker Position Reconstruction}\label{sec:reconstruction}
\par The output of the segmentation in Section~\ref{sec:segmentation} are planar positions of the markers in both image planes. To determine their 3D position with respect to a world frame $\{w\}$, we order and reconstruct these planar points using the epipolar geometry of the biplane setup. 
\par To begin the reconstruction, a primary image frame (either AP or lateral) is chosen which will be used to order the marker points. A good choice of primary image frame is the one which has more marker coverage by area, thus further spaced markers. In this image frame, the intermediate points are sorted by Euclidean distance from the tip marker, until all markers have been sorted. If $\{f\}$ is the primary image view, the sorted list of marker positions is denoted ${^{f}\tilde{\mb{p}}_{int}}$. This sorting is included in Alg.~\ref{alg:estimation} in Line 3 as the $\texttt{PointSorting}$ function.
\par The corresponding points in the secondary image frame are found by aligning the closest points to the epipolar lines created by the ordered set of marker positions in the primary view. These epipolar constraints allow us to determine the optimal order of the marker positions from the secondary set. Continuing with the example of $\{f\}$ as the primary image view, the sorted list of marker positions in the secondary image frame is denoted ${^{s}\tilde{\mb{p}}_{int}}$. If markers are missing, due to occlusion or overlapping markers in an image plane, these markers are excluded from the optimization problem. The index of these missing markers are found by comparing the spacing of consecutive markers and the marker design (see Section \ref{sec:missingsimulation}). With both sets of marker positions ordered, the 3D vector of marker positions can be found by using the projection matrices of each image plane with respect to the world:
\begin{equation}
    {^{w}\mb{p}_{i}} = \begin{bmatrix}
        {^{w}\mb{P}_{f}} \\ {^{w}\mb{P}_{s}}
    \end{bmatrix}^{+}\begin{bmatrix}
        \Rot{w}{f}{^{f}\mb{p}_{i}} \\ \Rot{w}{s}{^{s}\mb{p}_{i}}
    \end{bmatrix}
\end{equation}
where $i$ refers to a marker index ($i\in\left[b',b,1,...,n,e\right]$), and ${^{w}\mb{P}_{f,s}}$ refers to the projection matrix onto the front or side image plane, which is defined: ${^{w}\mb{P}_{f,s}} = \left[\mb{I} - \hat{\mb{n}}_{f,s}\hat{\mb{n}}_{f,s}\T\right]$,
%
%
where $\hat{\mb{n}}_f$ and $\hat{\mb{n}}_s$ refer to unit normal vectors of the front and side imaging planes. The output of this reconstruction, referred to as $\texttt{EpipolarReconstruct}$ on Line 4 of Alg.~\ref{alg:estimation}, is an ordered set of markers defined with respect to a world coordinate frame:
\begin{equation}\label{eq:array_positions_world}
    {^{w}\tilde{\mb{p}}} = 
        \left[{^{w}\mb{p}_{b'}},  {^{w}\mb{p}_{b}}, {^{w}\mb{p}_{1}}, \dots, {^{w}\mb{p}_{n}}, {^{w}\mb{p}_{e}}\right]
     \in\realfield{\left(3\times n+3\right)}
\end{equation}
%
%
%
\subsection{Solving for Modal Coefficients}\label{sec:shapesolve}
\par The ordered and reconstructed markers are used to estimate the shape of the catheter segment by finding the modal coefficients $\mb{c}$ that best match the forward kinematics to each marker point of this reconstructed point cloud. Note that we still do not know the base rotation angle $\sigma$. While theoretically one can solve for both $\mb{c}$ and $\sigma$ as part of a nonlinear optimization scheme, in practice the two are coupled and we solve the problem using a sequential process described in Section~\ref{sec:seqOpt}.
%
\par To start, assuming $\sigma=0$ we have the base frame origin as defined in Eq.~\eqref{eq:array_positions_world} (${^{w}\mb{p}_{b}}$). The orientation of the backbone curve local tangent at the base frame (${}^w\uvec{z}_b$) must also be estimated by defining the vector between the two base points,${^{w}\hat{\mb{z}}_{b}} = \left[{^{w}\mb{p}_{b}}-{^{w}\mb{p}_{b'}}\right]/{\|{^{w}\mb{p}_{b}}-{^{w}\mb{p}_{b'}}\|}$, which leaves the only unestimated degree-of-freedom (DoF) as the rotation about this axis, $\sigma$, which is solved in Section~\ref{sec:posesolve}.
\par Given an estimate of the base frame, ${^{w}\mb{T}_{b}} = \mb{T}(0)$, we solve for the optimal coefficients, $\mb{c}^{*}$, which minimize the sum squared-error of the modeled marker positions to their measured positions. This optimization problem, termed $\texttt{SolveShape}$ in Alg.~\ref{alg:estimation}, is defined:
\begin{equation}\label{eq:solveshape}
\begin{aligned}
 \min_{\mb{c}^{*},\sigma^{*}} \quad & \frac{1}{2}\Big(\breve{\mb{e}}(\mb{c},\sigma)\T\mb{W}\breve{\mb{e}}(\mb{c},\sigma) + \zeta_{c}\left(\mb{c} - \bar{\mb{c}}\right)\T \left(\mb{c} - \bar{\mb{c}}\right)+\\
 &\zeta_\sigma(\sigma-\bar{\sigma})^2\Big)\\
\textrm{where} &\quad  \breve{\mb{e}}(\mb{c},\sigma) = \left[\mb{e}_{1}(\mb{c},\sigma)\T,\dots,\mb{e}_{n}(\mb{c},\sigma)\T,\mb{e}_{e}(\mb{c})\T\right]\T\\
 \quad & \mb{e}_{i}(\mb{c},\sigma) = {^{w}\mb{p}_{i}}-\mb{p}_{i}(s_{i},\mb{c},\sigma), \quad i=1\ldots,n\\
 \quad & \mb{e}_{e}(\mb{c}) = {^{w}\mb{p}_{e}}-\mb{p}(L,\mb{c})
\end{aligned}
\end{equation}
where $\mb{W}$ is a diagonal weight matrix with all intermediate markers weighed similarly, and a higher weighted tip error ($\mb{e}_{e}(\mb{c})$). The second and third cost terms, with $\bar{\mb{c}}$ and $\left(\bar{\sigma}\right)$ referring to a moving average of previous shapes and roll, are damping terms to minimize sensitivity to noise, where $\zeta_{c}$ and $\zeta_{\sigma}$ are damping factors. The $\sigma$ term is included in the positions of the intermediate markers via:
\begin{equation}
    \mb{p}_{i}(s_i,\mb{c},\sigma) = \mb{p}(s_{i},\mb{c}) + \begin{bmatrix}
        r\cos(\beta_{i} + \sigma) \\ r\sin(\beta_{i} + \sigma) \\ 0
    \end{bmatrix}.
\end{equation}
\subsection{Solving for rotation about the backbone}\label{sec:posesolve}
\par With the shape of the central backbone known, the only remaining DoF needed is the roll angle, $\sigma$, about the local tangent vectors along the backbone. By solving this single angle the entire pose and shape is known along the entire length of the tracked catheter section.
%
%
\par This form of the optimization is referred to as $\texttt{SolveRoll}$ in Alg.~\ref{alg:estimation}. Both forms of Eq.~\eqref{eq:solveshape} may be solved using nonlinear least-squares. In this work, we used MATLAB's $\texttt{lsqnonlin}()$ function.
\subsection{Sequential Optimization}\label{sec:seqOpt}
\par The optimizations described in the preceding sections are solved serially because the shape and roll angle are inherently coupled if the modal coefficients define strain in a body-frame notation. For example, for any solution ($\mb{c}^{*},~\sigma^{*}$) there can be infinitely many alternative solutions which satisfy the same shape and pose. Therefore, to ensure the solution for shape and pose are optimal, these two optimizations should be solved iteratively until convergence of an optimal solution is achieved, as shown in Lines 5-11 in Alg.~\ref{alg:estimation}.
%
\begin{algorithm}
\caption{Shape and Pose Estimation Overview}\label{alg:estimation}
\begin{algorithmic}[1]
\Require $\text{img}_{f}$, $\text{img}_{s}$, $^{w}\mb{R}_{f,s}$, $\mb{c}_{0}$, $\sigma_{0}$, $\epsilon_{\mb{c}}$, $\epsilon_{\sigma}$
\State $^{f}\tilde{\mb{p}}_{u}$ = ImageSegmentation($\text{img}_{f}$)
\State $^{s}\tilde{\mb{p}}_{u}$ = ImageSegmentation($\text{img}_{s}$)
\State [$^{f}\tilde{\mb{p}}$,~$^{s}\tilde{\mb{p}}$] = PointSorting($^{f}\tilde{\mb{p}}_{u}$,~$^{s}\tilde{\mb{p}}_{u}$)
\State $^{w}\tilde{\mb{p}}$ = EpipolarReconstruct($^{f}\tilde{\mb{p}}$,~$^{s}\tilde{\mb{p}}$,~$^{w}\mb{R}_{f,s}$)
\While {$\Delta\mb{c}>\epsilon_{\mb{c}}~\textbf{or}~\Delta\sigma>\epsilon_{\sigma}$}
    \State $\mb{c}^{*}$ = SolveShape($^{w}\tilde{\mb{p}}$, $\mb{c}_{0}$,$\sigma_0$)
    \State $\sigma^{*}$ = SolveRoll($^{w}\tilde{\mb{p}}$, $\mb{c}^{*}$)
    \State $\Delta\mb{c} = \lVert\mb{c}^{*} - \mb{c}_{0}\rVert$
    \State $\Delta\sigma = \|\sigma^{*} - \sigma_{0}\|$
    \State $\mb{c}_{0} \gets \mb{c}^{*}$
    \State $\sigma_{0} \gets \sigma^{*}$
\EndWhile
\end{algorithmic}
\end{algorithm}
%

\section{Design Optimization}\label{sec:designOpt}
\begin{figure*}[htbp]
        \centering
        \includegraphics[width=0.85\textwidth]{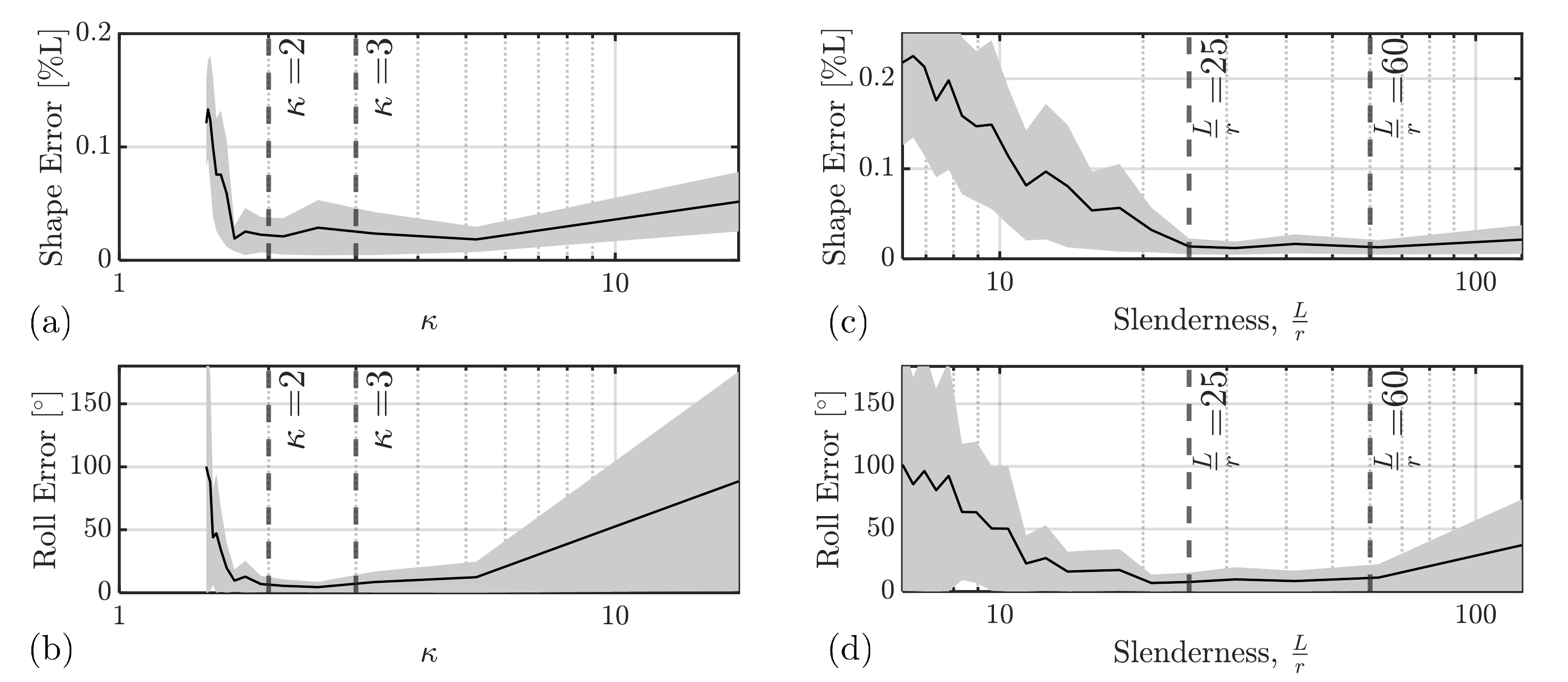}
        \caption{Simulation design study showing the error of the estimated shape as a function of marker spacing factor $\kappa$ in (a,b) and as a function of catheter slenderness, $L/r$ in (c,d). (a,c) RMS position errors (normalized by $L$), (b,d) RMS roll errors.  Solid lines denote the RMS error curves. Shaded regions visualize the confidence intervals of $\pm1$ standard deviation. Dashed lines show lower and upper bounds of recommended design regions.}\label{fig:design_study_results}
\end{figure*}
\par For any arbitrary catheter, the placement of the intermediate markers should be chosen to maximize the shape and pose estimation accuracy, and minimize the estimation sensitivity to uncertainty, such as segmentation noise. While the catheter radius, $r$, is specific to the catheter, the parameters which have design freedom include the tracked segment length, $L$, the number of radial markers, $n$, their arc-length placement, $\mb{s}_d$, and circumferential angle, $\bs{\beta}_d$. Although the formulation described in Section~\ref{sec:estimation_helical_kinematics} allows an arbitrary arrangement of markers, we will assume that the markers lie equally spaced on a helical path to reduce our design variables from $2n$ to two design variables: the helix pitch, $\lambda$, and the angular spacing  $\Delta\beta$ between markers. Furthermore, while it is assumed that the tracked length is provided, the slenderness of the catheter, defined by the ratio $L/r$, and the possibility of occluded markers also affect the performance of this estimation method.
\par The choice of marker design balances having sufficient markers to ensure robust shape tracking with sufficient marker spacing to be robust to segmentation uncertainty. For a straight helix, the spacing distance between two subsequent markers separated by a spacing angle $\Delta \beta$ is given by:
\begin{equation}
    d = \sqrt{4r^{2}\sin^{2}{\left(\frac{\Delta \beta}{2}\right)} + \lambda^{2} \Delta \beta^{2}}
\end{equation}
where the first term refers to the square distance of the chord between markers in a cross-section of the catheter, and the second term is the squared distance along the length of the catheter axis.
\par The segmentation uncertainty in either image plane, denoted as $\mathcal{N}_{f}$ and $\mathcal{N}_{s}$, provides the overall marker position uncertainty $\mathcal{N}$ due to segmentation. This uncertainty increases with the size of the markers, which can provide the designer with an initial estimate of expected uncertainty limits. For example, the centroid of a 2mm diameter marker will have more uncertainty than that of a 1mm diameter marker.
\par An optimal marker design for a given catheter will have a marker spacing, $d$, that is larger than the marker segmentation uncertainty $\mathcal{N}$. The design measure, which quantifies this relative value will be denoted $\kappa$ and referred to as the~\textit{marker spacing factor}, $\kappa \triangleq d/\mathcal{N}$.
%
%
\subsection{Simulation Study: Fixed Marker Spacing Design}\label{sec:evalsimulation}
\par The catheter was simulated at 25 randomized configurations throughout its workspace for 15 marker placement alternatives, with a minimum of $n=1,~(\kappa\sim18)$ and maximum of $n=71,~(\kappa\sim1.5)$ equally spaced intermediate markers used. At each configuration, random noise within a maximum uncertainty, $\mathcal{N}_{f,s} = 0.5\textrm{mm}$, was applied to each marker in both imaging planes. These randomized projected points in either plane were used to simulate the approach of Section~\ref{sec:estimationapproach}. The shape error, quantified using the root mean squared (RMS) positional error $e_p$, is defined as:
\begin{equation}\label{eq:shape_error}
    e_p=\sqrt{\frac{1}{k} \sum_{i=1}^{k}\Delta\mb{p}_i^2 }
\end{equation}
where $\Delta\mb{p}_i$ denotes the position error of the $i^{th}$ frame along the backbone, $k$ is the number of discrete frames along the segment, and $e_r$ the roll error between the estimated and actual rotation along the simulated backbone.
%
%
%
\par Figure~\ref{fig:design_study_results}(a-b) presents plots of these position and orientation errors. The vertical dashed lines denote the recommended minimum and maximum marker spacing, for this study found as $2<\kappa <3$. This result is interpreted as the spacing of consecutive markers should be at least twice the radius of the marker position uncertainty, however much higher spacing will minimize the signal available for shape and pose tracking.
\subsection{Simulation Study: Catheter Slenderness}\label{sec:slendersimulation}
\par A similar study was performed, evaluating the effect of changes in the catheter's slenderness ($L/r$). For a given catheter length of 25mm, varying radii from $r=0.2$mm to $r=4$mm were used to create 20 catheter designs, each with 25 equally spaced markers. These 20 designs were simulated in 25 random configurations with uncertainty of $\mathcal{N}_{f,s} = 0.5\textrm{mm}$.
\par In Fig.~\ref{fig:design_study_results}(c-d), the shape and roll error are reported, with low slenderness conceding poor shape information, but with significantly high slenderness showing an increase in roll error. The region between $25<L/r<60$ provides the most favorable results for estimation. For a 25mm long catheter section, this would indicate good performance for catheters with radius $0.4\textrm{mm}<r<1\textrm{mm}$. 
\begin{figure*}[htbp]
        \centering
        \includegraphics[width=0.95\textwidth]{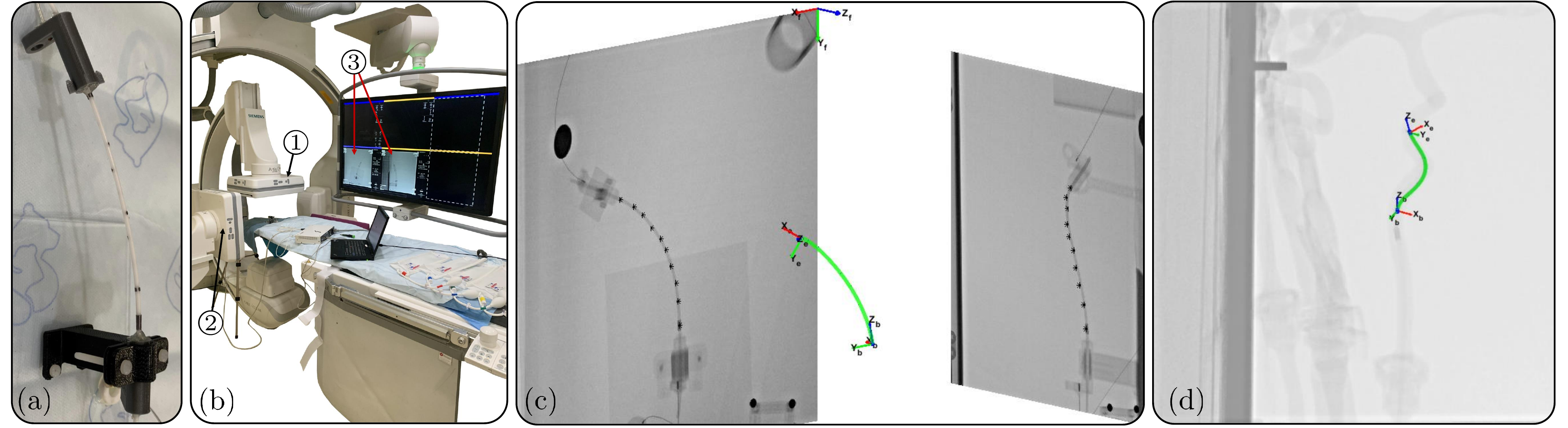}
        \caption{Experimental Setup in Operating Room. (a) The Tecothane prototype fixed on the patient bed. (b) shows the setup under the biplane fluoroscopy setup, (c) shows a sample estimation reconstructed from biplanes. (d) shows the estimated result overlayed on the Lateral plane. In (b), the circled numbers refer to the following: \protect\circled{1}-\protect\circled{2} are the AP and lateral imagers, respectively. \protect\circled{3} show the AP and lateral displays present to the interventionalist during a standard procedure.
        }\label{fig:experimental_setup}
\end{figure*}
\par These results may indicate to a steerable catheter designer how long of a distal section may be tracked, and how to orient the intermediate marker for favorable catheter tracking performance with this method.
\subsection{Simulation Study: Dropped Markers}\label{sec:missingsimulation}
\par To evaluate the robustness of this method to marker occlusion, we repeated this simulation for a single marker design across the same 25 configurations. At each configuration, at least one marker and up to half of the markers were randomly excluded from one image view. The algorithm sorted point correspondence and determined which markers were missing, and evaluated the shape and pose. The shape and pose estimate with all markers included was recorded as a control. Both control and dropped marker results are included in Table~\ref{tab:dropped_marker_errors}. These show that even with up to half of the markers missing for this given design, the algorithm can still return considerably accurate shape and roll information.
\begin{table}[htbp]
\scriptsize
\centering
\begin{tabular}{|c|c|c|c|c|c|}
\hline
\begin{tabular}[c]{@{}c@{}}Estimation\\ Order ($m$)\end{tabular} & \begin{tabular}[c]{@{}c@{}}Shape Error\\Control\end{tabular} & \begin{tabular}[c]{@{}c@{}} Shape Error\\Dropped\end{tabular} & \begin{tabular}[c]{@{}c@{}}Roll Error\\Control\end{tabular} & \begin{tabular}[c]{@{}c@{}}Roll Error\\Dropped\end{tabular} \\ \hline
2 & 3.34\% & 5.88\% & $27.71^{\circ}$ & $45.34^{\circ}$\\ \hline
3 & 1.42\% & 3.92\% & $7.96^{\circ}$ & $21.04^{\circ}$\\ \hline
4 & 1.47\% & 4.58\% & $10.94^{\circ}$ & $24.72^{\circ}$\\ \hline
\end{tabular}
\caption{RMS Errors of shape and roll for the control (no dropped markers) and the dropped marker cases. Shape errors reported as percentage of segment length.}\label{tab:dropped_marker_errors}
\end{table}

%
\section{Evaluation} \label{sec:eval}
%
%
%
\subsection{Experimental Setup}\label{sec:exp_setup}
\subsubsection{Workspace Configurations}
\par The experimental validation of this approach involved collecting biplane fluoroscopy images using the Artis Q system (Siemens Healthineers, Erlangen, Germany). The prototype used was a \diameter2mm (0.079") Tecothane\texttrademark ~tube which is flexible in bending, but relatively stiff in torsion, shown in Figure~\ref{fig:catheter_frames}(d).
Along the surface of the prototype, 0.025mm (0.001") thick Tantalum foil (Thermo Fisher Scientific, Waltham, MA, USA) bands were placed at the tip and base, and circular markers were placed at known surface locations between the bands. Tip and base bands were 2mm wide, while circular markers were 1.5mm in diameter.
%
\par The prototype was imaged under the biplane fluoroscopy system (see Fig.~\ref{fig:experimental_setup}), and was oriented statically in different poses and shapes. At each configuration, a still from a stream of fluoroscopy or a static X-ray image was collected. Ground truth locations were manually selected from each corresponding image. The DICOM files storing each image also provided imaging calibration parameters including the pixel resolution (<0.35mm) and the orientation of the imaging plane. With each image scaled and oriented, the shape and pose estimation was performed following the approach described throughout Section~\ref{sec:estimationapproach}. This process was repeated for a range of configurations of varying bending and base rotation.
\subsubsection{In-Vitro Insertion}
\par An additional prototype was made with similar markers, however placed on standard \diameter1.4mm (0.055") microcatheter (Phenom Plus, Medtronic). This prototype was inserted in the left vertebral artery of a silicone phantom model (Neuro System Trainer, United Biologics), as shown is shown in Fig.~\ref{fig:experimental_setup}(d).
\subsection{Experimental Results}\label{sec:exp_results}
\subsubsection{Workspace Configurations}
\par The tracking results for 29 configurations of the catheter are included in Table~\ref{tab:experimental_errors}. For each configuration, the shape and pose were estimated with Chebyshev polynomials of $m=2,3,4$. One configuration's estimation is reported in Figure~\ref{fig:experimental_setup}(c), and is shown in the multimedia extension. The positional shape error is computed by finding the Euclidean distance between the estimated marker positions, and the reconstructed ground truth locations selected from each image frame. The orientation error for the catheter pose is decomposed into error in the local tangent direction, and roll error about the backbone. The tip direction (tangent) errors were computed at the base frame, $\{b\}$, and tip frame, $\{e\}$. The positional and tangent errors in Table~\ref{tab:experimental_errors} indicate high accuracy in computing the shape of the catheter segment, with positional errors $<2\%$ of the catheter's length and tangent errors of $<12^{\circ}$. The roll error was computed by finding the angular error between the estimated and ground truth marker locations within the x-y plane of the estimated frame. While the reported roll error is higher ($\sim30^{\circ}$) than the tangent errors, since roll is more sensitive than shape to uncertainty at such small catheter diameters, it is noted that this information within $45^{\circ}$ may still be useful to interventionalists who have no estimate of roll otherwise.
\par We note that the experiments described above are considered a worst-case scenario in terms of convergence. In a practical application, the motion of the catheter will be tracked continuously and therefore the algorithm may be initialized with a greatly improved initial guess, while in these estimations we used an initial guess of $\mb{c}_0 = \mb{0}$ and $\sigma_0=0$. 
\begin{table}[htbp]
\scriptsize
\centering
\begin{tabular}{|c|c|c|c|c|}
\hline
\begin{tabular}[c]{@{}c@{}}Estimation\\ Order ($m$)\end{tabular} & \begin{tabular}[c]{@{}c@{}}RMS Shape Error\\ (Percent Length)\end{tabular} & \begin{tabular}[c]{@{}c@{}}RMS Tangent\\ Error (SD)\end{tabular} & \begin{tabular}[c]{@{}c@{}}RMS Roll\\ Error (SD)\end{tabular} \\ \hline
2 & 0.98mm (1.25\%) & $6.60^{\circ}~(3.03^{\circ})$ & $32.22^{\circ}~(12.38^{\circ})$\\ \hline
3 & 0.92mm (1.18\%) & $8.21^{\circ}~(4.54^{\circ})$ & $30.54^{\circ}~(13.74^{\circ})$\\ \hline
4 & 0.89mm (1.14\%) & $11.14^{\circ}~(7.28^{\circ})$ & $28.90^{\circ}~(12.99^{\circ})$\\ \hline
\end{tabular}
\caption{Experimental Errors for different approximation orders. Note the shape errors are defined as true value (in mm) and also as a percentage of the segment length ($\%$L).}\label{tab:experimental_errors}
\end{table}
\subsubsection{In-Vitro Insertion}
The insertion included 134 fluoroscopy frame sequences as the catheter navigated the left vertebral artery. Across these frames, with an modal order of 3, the RMS shape and roll error was found to be 0.896mm (2.62\%) and $38.8^{\circ}$, respectively. The minimal curvature during this navigation is 2.50mm, which is tighter than the average radius of curvature of the carotid siphon \cite{zhang2013siphon}. Future work will incorporate time-series history with an Extended Kalman Filter to improve the estimation under uncertainty. Additional sources of uncertainty include the measurement of marker locations, as the $\beta$ terms could only be measured within $\sim5^{\circ}$.
%

%
\section{Conclusion}\label{sec:conclusion}
\par In this paper, we demonstrate feasibility of an image-based shape and pose tracking algorithm for catheters under biplane fluoroscopy. Where existing works determine either shape sensing assuming knowledge along the full length of the catheter, or pose-tracking using 6-DoF sensors at discrete locations, this work provides both information from fluoroscopic imaging alone. This tracking relies on a catheter equipped with radiopaque markers along the catheter's outer surface, whose position along the segment arc-length and rotation about the central axis are known a-priori. These markers are then segmented in either imaging plane and reconstructed in the world frame. Given these reconstructed marker locations, the modal coefficients are solved to provide the catheter shape, and finally the rotation about the central axis is solved to provided an estimate of the 6-DoF pose at every point along the catheter's distal segment. This sensing algorithm was deployed for two $\leq\diameter2\textrm{mm}$ catheter prototypes under biplane fluoroscopy, including navigation into a phantom vertebral artery. By tracking both shape and pose, navigation of steerable catheters can be simplified and made safer for interventionalists navigating tortuous structures of the brain vasculature.

\bibliographystyle{IEEEtran}
\bibliography{bib/shape_estimation}

\balance
\end{document}